\documentclass[a4paper]{article}

\usepackage{geometry}
\usepackage{svg}
\usepackage{subfigure}
\usepackage{graphicx}
\usepackage{hyperref}
\usepackage{xcolor}
\usepackage{booktabs}
\usepackage{algorithm}
\usepackage{algpseudocode}
\usepackage{amsthm}
\theoremstyle{plain}
\usepackage{amsfonts}
\usepackage{amssymb}
\usepackage{amsmath}
\usepackage{graphicx}
\usepackage{psfrag}
\usepackage{xcolor}
\usepackage{enumerate}
\usepackage{lineno}
\usepackage{mathtools}
\usepackage{tikz-cd}
\usepackage[labelfont=bf]{caption}

\usepackage[inline]{enumitem}
\usepackage[title]{appendix}
\usetikzlibrary{arrows}
\usepackage{cleveref}

\usepackage[all]{xy}
\newtheorem*{theorem*}{Theorem}

\newtheorem*{proposition*}{Proposition}
\newtheorem*{lemma*}{Lemma}

\newtheorem{remark}{Remark}
\newcommand{\R}{{\mathbb{R}}}
\newcommand{\Rl}{{(\mathbb R, \le)}}

\begin{document}

\title{Persistence-based operators in machine learning}
\author{Mattia G. Bergomi\\
Independent Researcher\\
Milan, Italy \\
\small{\texttt{mattiagbergomi@gmail.com}} \\
\and
Massimo Ferri\\
ARCES and Dept of Mathematics,\\
University of Bologna, Italy\\
\small{\texttt{massimo.ferri@unibo.it}}\\
\and
Alessandro Mella\\
Independent Researcher\\
Verona, Italy\\
\small{\texttt{alessandro.mella.92@gmail.com}}\\
\and
Pietro Vertechi\\
Independent Researcher\\
Trieste, Italy\\
\small{\texttt{pietro.vertechi@protonmail.com}}}

\maketitle
\begin{abstract}
  Artificial neural networks can learn complex, salient data features to achieve a given task. On the opposite end of the spectrum, mathematically grounded methods such as topological data analysis allow users to design analysis pipelines fully aware of data constraints and symmetries. We introduce a class of \textit{persistence-based} neural network layers. Persistence-based layers allow the users to easily inject knowledge about symmetries (equivariance) respected by the data, are equipped with learnable weights, and can be composed with state-of-the-art neural architectures.
\end{abstract}

\section{Introduction}

Topological data analysis and topological persistence (TP) rely on user-defined functions to generate concise fingerprints of topological objects, e.g., triangulable manifolds and simplicial complexes. These fingerprints carry information about the geometrical and topological features of an object as seen through the lenses of the user-defined function. On the other hand, artificial neural architecture layers learn weights to represent their input optimally for a user-defined task. 

TP requires users to endow the data with a topological structure and then devise specialized continuous functions to highlight salient features of the transformed dataset for the task at hand. Artificial neural networks learn such functions through gradient optimization of a differentiable error function and provide layer templates adapted to the structure of most data types, e.g., convolutional layers~\cite{lecun1999object} for images and recurrent architectures~\cite{yu2019review} for time-varying signals. However, only few theoretical frameworks and algorithms provide ways to inject knowledge into neural networks.

Interactions between TP and deep learning are of broad interest~\cite{cang2018integration,bergomi2019towards,pun2022persistent}. Ideally, combining the two methods would yield easily constrainable and learnable layers, composable with state-of-the-art neural network layers. We believe that the need to map data to topological objects and express their features as critical points of continuous functions hinders the development of TP-inspired neural layers.

\paragraph*{Aim.} We provide a plain approach to the design of persistence-based layers for artificial neural architectures. Particularly, we leverage rank-based persistence---a generalization of topological persistence~\cite{bergomi_rank-based_2020}---to avoid complex auxiliary constructions mapping data to topological objects and work, when needed, with more natural data representations. Then, thanks to the notion of persistent features developed in~\cite{bergomi2022steady}, we define persistence-based operators taking advantage of the properties of the data itself. 

\paragraph*{Contribution.} We provide a streamlined introduction to topological and rank-based persistence, and to persistent features. Then, we define and give constructive examples of operators based on persistent  features and discuss their properties, particularly equivariance (constrainability) and noise robustness. After showcasing these properties on images,  we define the persistent feature-based neural network layer. We base our proposal of persistence-based layer on two main principles. On the one hand, we aim to learn relevant (persistent) features from data points. On the other hand, we provide simple strategies to take advantage of locality and equivariance, two key notions for the analysis of structured data.
We provide an algorithm for a persistence-based pooling operator, test it on several architectures and datasets, and compare it with some classical pooling layers.

\paragraph*{Structure.} \Cref{sec:background} provides intuition on the central mathematical concepts involved in the definition of persistence-based layer: topological persistence (e.g., persistent homology), rank-based persistence, and persistent feature. In~\cref{sec:persistence_based_layers}, we devise an image filtering algorithm based on persistent features, discuss its main properties, and provide examples. Thereafter, we define a persistence-based neural network layer and specialize it to act as a pooling layer in convolutional neural networks. Computational experiments in~\cref{sec:computational_experiments} evaluate and compare the performance of the proposed pooling layer in an image-classification task on several datasets and on two different architectures. In the same section, we provide a qualitative analysis of the most salient features detected by the persistence-based pooling, the classical max-pooling, and LEAP~\cite{sun2017learning}.

\section{Background}\label{sec:background}

Topological persistence and persistent homology have been extended in several ways, e.g.,~\cite{bubenik_categorification_2014,Les15,oudot2015persistence,mccleary2020bottleneck,govc2021complexes}. However, we place our proposal in the rank-based framework to work as independently as possible from auxiliary mappings from the data points to topological spaces.

In the following paragraphs, we intuitively define and provide references for the essential mathematical constructions that motivate and inspire the idea of persistent-based layers: persistent homology, rank-based persistence, and steady persistent features.

\subsection{Rank-based persistence}\label{sec:classical_to_general}

\begin{figure}[tb]
  \centering
  \includegraphics[width=.9\textwidth]{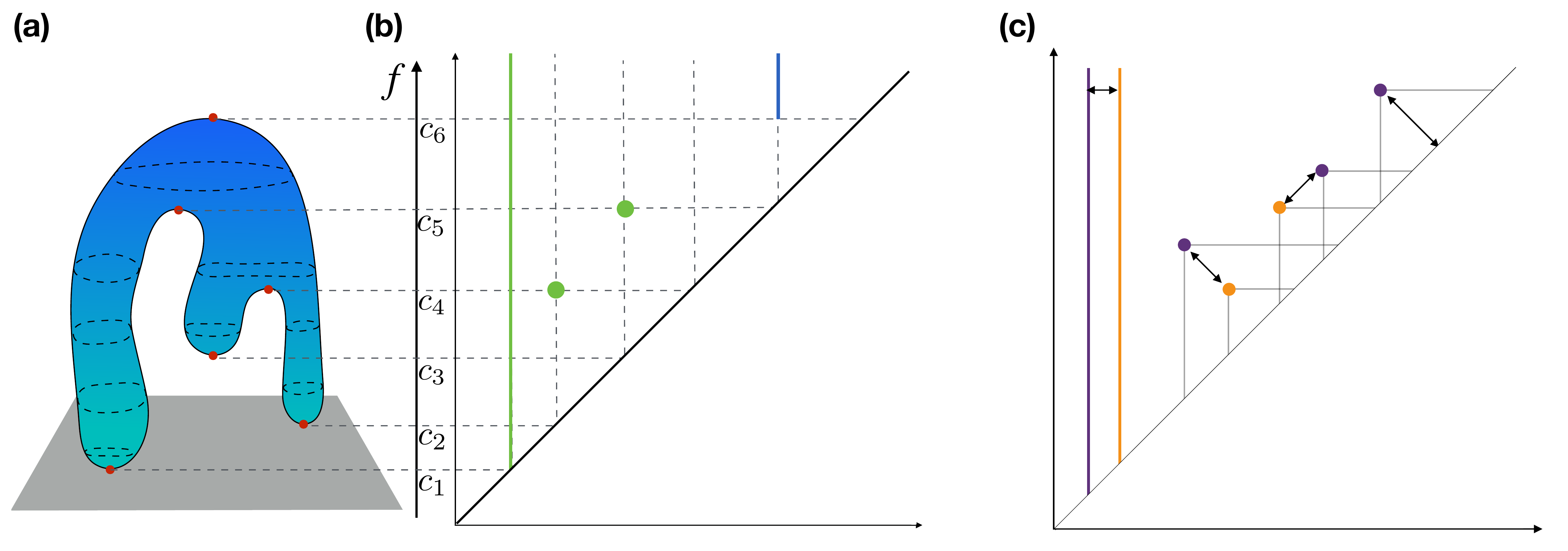}
  \caption{\textbf{Persistent homology}. Let us consider a topological space $X$ and a continuous function $f:X\rightarrow\R$. The (homological) critical values of $f$ induce a sub-level set filtration of $X$. \textbf{(a)} Critical values $C = \{c_1, \dots, c_6\}$ of the height function on a a topological sphere. Sublevels of the filtering function $f$---namely $f^-1\left((-\infty, c]\right)$, for every $c\in C$---yield a filtration of the topological sphere. \textbf{(b)} Changes in number of generators of the $k$th homology groups along the filtration can be represented as a persistence diagram. $0$-dimensional holes (connected components) are represented as green points. The void obtained at the last sublevel set gives rise to the blue line. \textbf{(c)} Persistence diagrams can be compared by computing an optimal matching of points. Unmatched points are associated with their projection on the diagonal. \label{fig:persistence_bkg}}
\end{figure}

Persistent homology requires three main ingredients:
\begin{enumerate*}
  \item A filtered topological space,
  \item the homology functor $S_k$ mapping topological spaces to finite vector spaces, and
  \item a notion of \textit{rank}, e.g., dimension for vector spaces or cardinality for sets~\cite{bergomi2021beyond}.
\end{enumerate*}
See~\cref{fig:persistence_bkg} for an exemplification of persistent homology, and~\cite{edelsbrunner_topological_2000} for details.

We choose to frame our work in a more general context than homological persistence, namely \textit{rank-based persistence}~\cite{bergomi_rank-based_2020}. There, assuming an axiomatic standpoint based on the three building blocks of homological persistence mentioned above, the authors generalize persistence to categories and functors other than topological spaces and homology. Importantly, under few assumptions, persistence built in the rank-based framework still guarantees fundamental properties such as flexibility (dependence on the filtering function), stability~\cite{cohen-steiner_stability_2007}, and robustness~\cite{di2011mayer}. In this setting, we can work with data types such as images and time series, without intermediary topological constructions. We refer to~\cite{bergomi_rank-based_2020} for details and provide a list of analogies between classical and rank-based persistence in~\cref{tab:analogies}.

\begin{table}[tb]
  \begin{center}
  \begin{tabular}{@{}ll@{}}\toprule
   \textbf{Classical framework} & \textbf{Categorical framework}\\\midrule
   Topological spaces & Arbitrary source category $\mathbf C$ \\
   Vector spaces & Regular target category $\mathbf R$\\
   Dimension & Rank function on $\mathbf R$\\
   Homology functor & Arbitrary functor from $\mathbf C$ to $\mathbf R$ \\
   Filtration of topological spaces & $\Rl$-indexed diagram in $\mathbf C$ \\
   \bottomrule
  \end{tabular}
  \caption{Analogy between the classical and rank-based persistence frameworks.\label{tab:analogies}}
  \end{center}
  \end{table}

  \begin{figure}[tb]
    \centering
    \includegraphics[width=\textwidth]{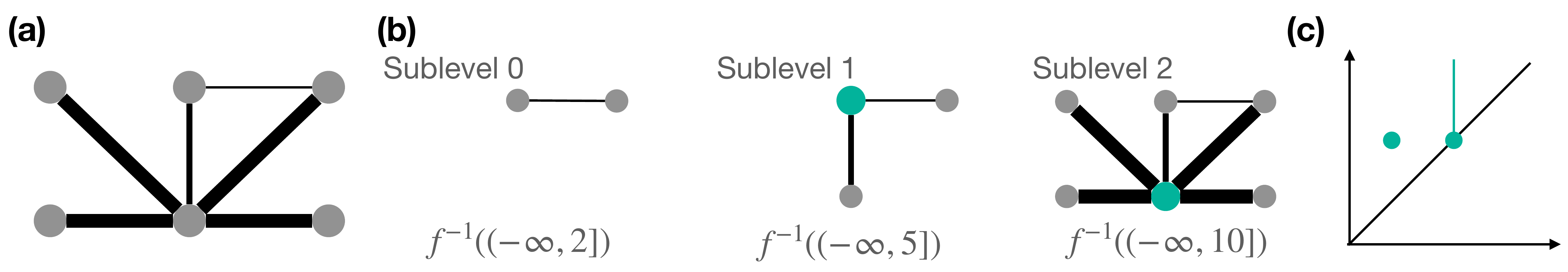}
    \caption{\textbf{Persistent features}. \textbf{(a)} A weighted graph $(G, f)$. The weight values---integers in the set $\{2, 5, 10\}$---are encoded as the thickness of the edges. \textbf{(b)} The sublevel set filtration induced by the weights. Teal vertices are the vertices in each sublevel with local degree prevalence. \textbf{(c)} The persistence diagram obtained by considering the steadiness of each true subset along the filtration. One vertex realizes local degree prevalence at level $5$ and stops being prevalent entering level $10$. Another vertex is prevalent at level $10$ and shall never stop being such because $f^{-1}((-\infty, 10])=G$. \label{fig:persistence_features}}
  \end{figure}

\subsection{Persistent features}
\label{sec:persistent_hubs}

\begin{figure}[tb]
    \centering
    \includegraphics[width=.8\textwidth]{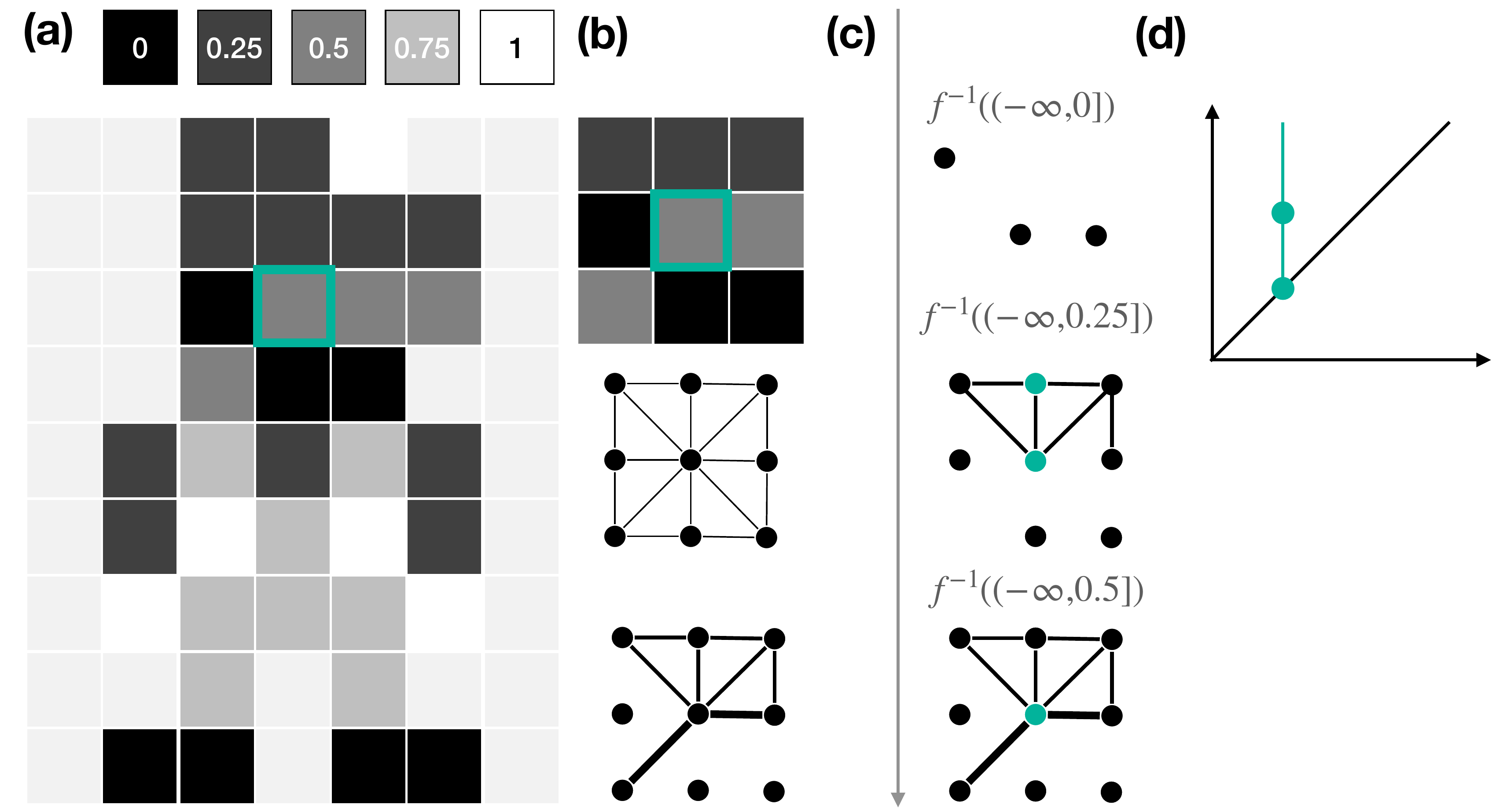}
    \caption{\textbf{Persistent features}. An image \textbf{(a)} and its interpretation as weighted graph. \textbf{(b)} Given a focus pixel and its neighborhood (a $3\times 3$ patch in the example), we map the image to a graph $G$ having as vertices the patch's pixels and edges connecting adjacent pixels. We weight edges of $G$ by evaluating the function $f:\R\times\R\rightarrow\R$ mapping the intensity values of two vertices to the minimum intensity values of the respective pixels in the image. Edges with weight $0$ are not showed. \textbf{(c)} The sublevel-set filtration induced by the weights. Teal vertices are the ones with prevalent local degree. \textbf{(d)} The persistence diagram obtained considering the filtration/feature pair. \label{fig:persistent_hubs}}
  \end{figure}

In the spirit of the aforementioned generalization, \cite[Sec. 2.2]{bergomi2022steady} introduces the concept of \textit{steady persistent features} for weighted graphs. 

A weighted graph is a pair $(G,f)$, where $G=(V,E)$ is a graph defined by a set of vertices $V$ and edges $E$, and $f$ is a function assigning (tuples of) real-valued weights to the edges of $G$, in symbols $f: e_i=[v_{i,1}, v_{i,2}] \mapsto w_i$. The weighting function naturally induces a sublevel set filtration
\[
    \emptyset\subseteq S_1 = f^{-1}((-\infty, w_1])\subseteq\cdots\subseteq S_n =f^{-1}((-\infty, w_n]) = G.
\]    
For an intuition, see~\cref{fig:persistence_features}, panels (a) and (b).

Let $S = 2^{V\cup E}$ be the set of all subsets of elements (vertices and edges) of $G$. Let $F$ be a graph-theoretical property, e.g., local degree prevalence, independence. The \textit{persistent feature} $\mathcal{F}:S\rightarrow \{\textrm{true}, \textrm{false}\}$ associated with $F$ is a boolean mapping returning true if the property $F$ holds for a certain subset $s\in S$ and false otherwise. Symmetrically to the topological persistence framework, we evaluate $\mathcal{F}$ for every subset of  every $\{S_i\}_i$. See~\cref{fig:persistence_features}, panel (b). Then, we compute the \textit{steadiness} $\sigma$ of each subset $s$ along the filtration by counting subsequent sublevels such that $\mathcal{F}(s) = \textrm{true}$. As in~\cite{bergomi2022steady}, we refer to $\sigma((G,f, \mathcal{F}))$ as the steady persistence of the feature $\mathcal{F}$ on $(G,f)$. This construction yields a persistence diagram. See~\cref{fig:persistence_features}, panel (c).

\section{Persistence-based layers}\label{sec:persistence_based_layers}

\begin{figure}[tb]
    \centering
    \includegraphics[width=.7\textwidth]{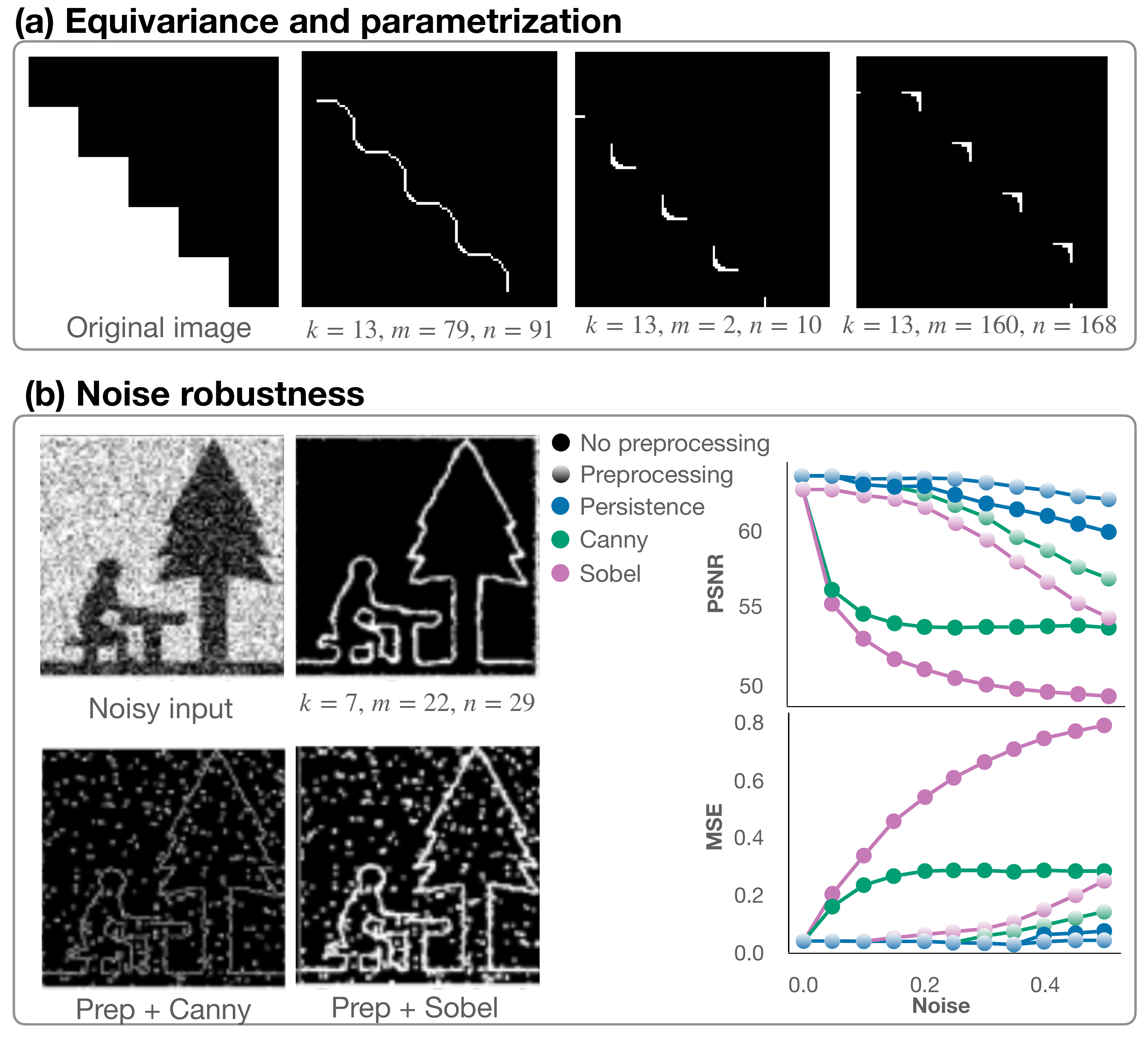}
    \caption{\textbf{Persistence-based filter}. \textbf{(a)}. The persistence-based filter acts in an equivariant fashion (the same feature is recognized throughout the image) and parametrization changes lead to detection of heterogeneous features. \textbf{(b)} We perform edge detection on a noisy input ($40\%$ of the pixels are salt and pepper noise). Performance are compared across the proposed persistence-based filter, Canny, and Sobel edge-detection algorithms with and without preprocessing (median filter). Results are showcased for different ratios of noisy pixels.\label{fig:filter}}
  \end{figure}

Applying the definition of steady persistent function, we build a persistence-based operator that can act as a filter on grayscale images (RGB images can be treated as by considering channels independently). We discuss important properties of such operators, such as locality and equivariance. Then, we adapt the operator to work as a pooling layer in a convolutional neural network.

\subsection{Persistent features as equivariant filters}\label{sec:filter}

Locality and equivariance are crucial features for convolutional neural networks, and in general for any group-equivariant model. Indeed,
\begin{enumerate}
    \item the intensity of a pixel in a grayscale image carries knowledge only when compared to neighboring pixels; 
    \item identical configurations located in different regions of the image (translated) are recognized by translation-equivariant models.
\end{enumerate}

\paragraph*{Locality.}
In this setting, considering the notion of persistent feature introduced in~\cref{sec:persistent_hubs}, we think about an image as a graph in which vertices are pixels and edges connect adjacent pixels. Patches (or windows) of size $k$ around a pixel (point) correspond to $k$-distance neighborhoods of the vertex associated to such pixel.

\paragraph*{Flexibility.} Persistent features require a filtered space to be computed. Thus, after associating a graph to an image (or time series), we define the filtration $f:S\rightarrow\R^n$, where $S$ is the set of all subsets of $V\cup E$ and $n\in\mathbb{N}$. Importantly, $f$ can carry additional information about the original data: when considering images, one can associate with each vertex the intensity value of its underlying pixel, and leverage this information to compute appropriate weights. See~\cref{fig:persistent_hubs}, panels (a) and (b). 

\paragraph*{Equivariance.} Once the pair $(G,f)$ has been associated to our data, the proposed construction is naturally equivariant with respect to translation. Indeed, sublevel set filtrations and persistent features are totally determined by the weights and connectivity of the graphs at each filtration level. It is important to notice that the flexibility of the proposed solution makes it possible to further control the equivariance of the operator. For instance, weighting functions that only depend on pairwise intensity values will generate operators that not only are equivariant with respect to translations, but also to isometries (translation, rotations, reflection, and scaling) of the original data.

\paragraph*{Parametrization.} We can create parametrized $f$ and $\mathcal{F}$ to create operators endowed with more complex equivariance and learnable parameters. 
Let $S_t = \iota^{-1}((-\infty, t])$ be a sublevel of $G$ naturally induced by the intensity of pixels (edges are added when the vertices they connect are added). We define $\mathcal{G}_{m,n}^k(v, t)$ as the feature mapping $v$ to true if more than $m$ and less than $n$ pixels in the $k$-distance neighborhood of $v$ have intensity less than $t$. Panel (a) of~\cref{fig:filter} shows how varying parameters $m$ and $n$ allows us to highlight radically different aspects of a binary image. 

\begin{remark}
    $\mathcal{G}$ does not rely on the $2$-dimensional structure of the selected patch, thus, its steady persistence diagram is not only invariant with respect to translations, but also to permutations. This kind of equivariance meshes the standard convolutional equivariance with the fully-connected input/output representation typical of dense layers of an artificial neural network.
\end{remark} 

\paragraph*{Robustness.} In the context of both topological and rank-based persistence, robustness means small changes in the image do not give rise to significant perturbations of the corresponding persistence diagram. This concept is defined in~\cite{bergomi2022steady} for persistent features and dubbed \textit{balancedness}. In~\cite{mella2021non}, it is formally shown that $\mathcal{G}$ is a \textit{balanced feature} in the sense of~\cite{bergomi2022steady}. Robustness to noise is tested in~\cref{fig:filter}, panel (b).

\subsection{A steady, persistent-feature layer}
\label{sec:pfl}

\begin{figure}[tb]
    \centering
    \includegraphics[width=\textwidth]{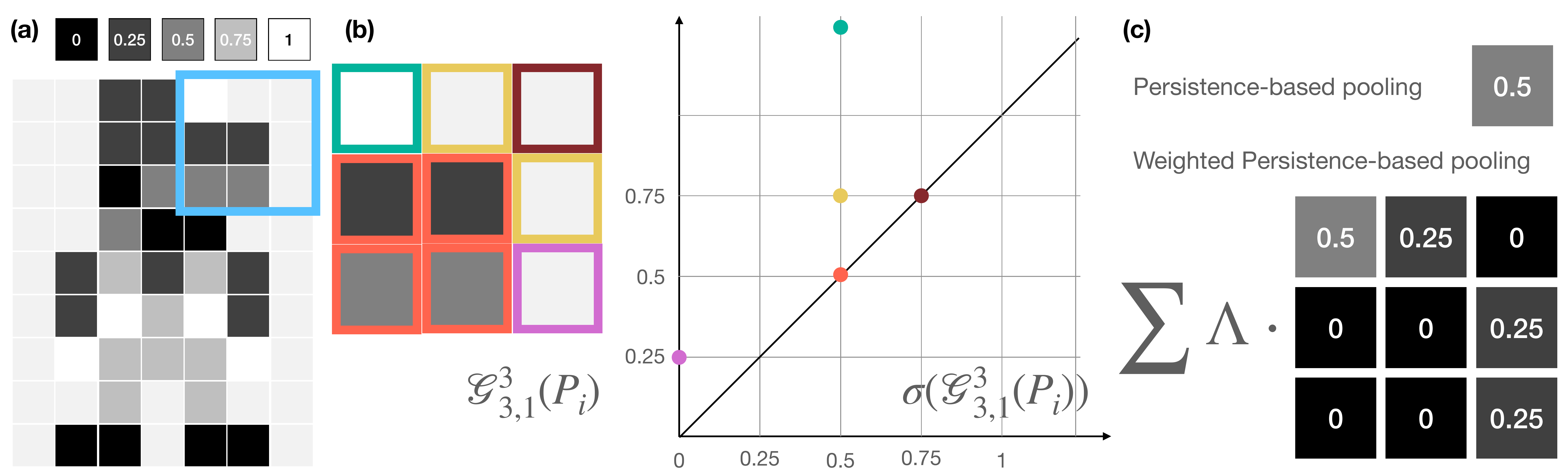}
    \caption{\textbf{Persistence-based Pooling}. \label{fig:pooling}}
  \end{figure}

The filter $\mathcal{G}$ and its steady persistence introduced in~\cref{sec:filter} are endowed with features inherited from their mathematical foundation which make them suitable for tackling typical machine learning tasks:
\begin{enumerate*}
    \item $\mathcal{G}$ enhances signal in correspondence of abrupt signal changes (a max-pooling filter could be blind to such features);
    \item the steady persistence $\sigma(\mathcal{G})$ yields equivariant representations of the input with respect to the group of isometry;
    \item salt-and-pepper noise does not impair the quality of detected features.
\end{enumerate*}

These properties motivate implementing and testing $\sigma(\mathcal{G})$ as a complement to pooling and convolutional operators in standard artificial neural networks. In the following paragraph, we discuss the computability of $\sigma(\mathcal{G})$, and then introduce some learnable, i.e., continuous, parameters allowing to fully integrate our operator in machine learning pipelines based on automatic differentiation. 

In the remainder, we discuss how to implement and add learnable parameters to the pooling case. The same operator can be easily adapted to work as a convolution-like layer.

\paragraph{Persistence-based pooling.} We consider an image $I$ and split it in a collection of patches $\{P(h, w)_i\}$ of size $(h, w)\in\mathbb{N}^2$. For every pixel $p\in P_i$, we compute $\sigma(\mathcal{G}_{m,n}^k)(p)$ for some fixed values of $m,n$ and $k$ (padding is added if needed). This procedure, which computationally boils down to sorting and slicing operations, yields a persistence diagram having a point per each pixel in $P_i$. Indeed, we compute the operator $\mathcal{G}$ for every of $P_i$, padding the patch whenever necessary. Symmetrically to the classical max-pooling operator, the maximum persistence---i.e., the distance from the diagonal---shall determine the value to be associated to the entire patch and its downsampling. As an example, see~\cref{fig:pooling}, panels (a), (b), and top row of (c).

\paragraph*{Parametrization.} Alternatively and following the idea of learnable pooling operators, e.g.~\cite{sun2017learning}, we propose to learn weights $\Lambda$ to modulate the contribution of each non-zero component of the persistence diagram, as depicted in~\cref{fig:pooling} panel (c). Because the persistence value associated with each pixel is continuous and $\mathcal{G}$ is balanced, we learn such weights through standard backpropagation.

\section{Computational experiments}
\label{sec:computational_experiments}

We assess the performance of the suggested pooling layer embedding it into two neural architectures. First, we compare the performance of the persistence-based pooling with some classical pooling implementations, then we test differences in detected saliency across the considered pooling layers.

\subsection{Datasets}
\label{sssec:Dataset}
We perform computational experiments on the MNIST~\cite{lecun-mnisthandwrittendigit-2010}, Fashion-MNIST~\cite{xiao2017online}, and CIFAR-10~\cite{krizhevsky2009learning} datasets. The MNIST and Fashion-MNIST datasets are composed of grayscale images labelled according to ten classes of hand-written digits and fashion articles, respectively. Both dataset are composed of 60000, 28 pixel by 28 pixel, black and white images for training, and 10000 for testing.
The CIFAR-10 dataset is composed of 50000, 32 pixel by 32 pixel, RGB images for training, belonging to ten classes: airplane, automobile, bird, cat, deer, dog, frog, horse, ship, and truck.  10000 labelled test images are also available.

\begin{figure}[tb]
  \centering
  \includegraphics[width=\linewidth]{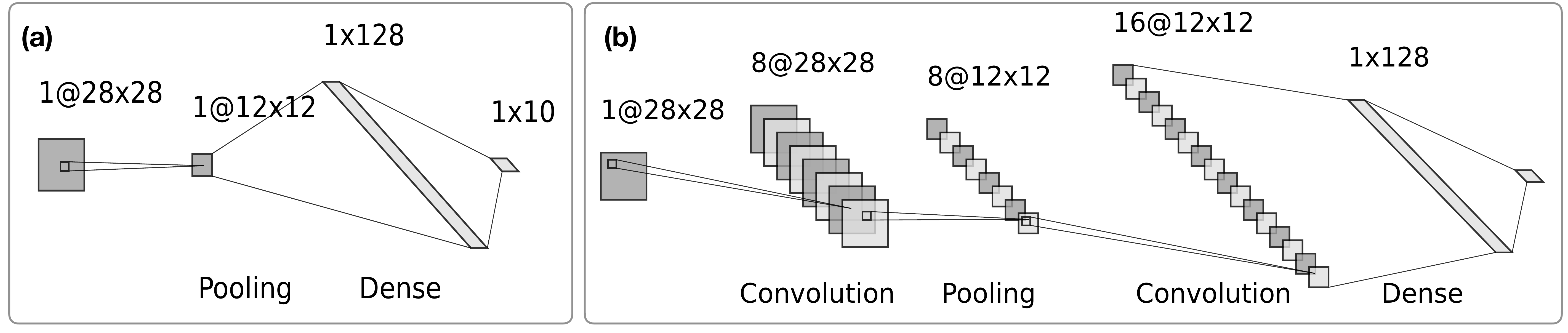}
  \caption{\textbf{Benchmark architectures.} The two architectures used in computational experiments. \textbf{(a)} A toy neural network where a pooling operator is applied directly to input images. The downsampled output of the pooling layer is passed to a dense classifier. In other words, first, the pooled output is flattened, then its dimensionality reduced to match the number of classes to be predicted by the network. \textbf{(b)} As common practice for CNNs, this architecture alternates convolutional and pooling layers before the dense classifier.}
  \label{fig:Arch}
\end{figure}

\subsection{Architectures}
\label{sec:architectures}
We test the proposed layer with two neural network architectures, see~\cref{fig:Arch}. We designed the first architecture to provide the simplest configuration in which different pooling operators could be compared during a supervised task where parameters are learned via gradient propagation. The second architecture is more akin to the standard convolutional neural network topology in a simple image-classification task: we alternate convolutional an pooling layers, before a dense classifier.

\paragraph*{Training.} We use sparse categorical cross-entropy as loss function~\cite{chollet2015keras}, and the ADAM optimizer~\cite{kingma2014adam} with learning rate $3e-4$. Images are fed to the networks in batches of size $32$. We accept a maximum of $100$ training epochs, setting early stopping~\cite{prechelt1998early} with parameters $\textrm{min\_delta} = 0$ and $\textrm{patience} = 3$.

\subsection{Results}
\begin{table}[tb]
  \begin{center}
\begin{tabular}{lllll}\toprule
             \textbf{Architecture} - \textbf{Dataset}     & \textbf{Max}    & \textbf{LEAP}   & \textbf{PL} (ours)     & \textbf{PML}  (ours)  \\\midrule
(a) - MNIST  & 0.8905 & 0.9238 & $\boldsymbol{0.9472}$ & 0.9087 \\
(b) - MNIST & 0.9886 & 0.9848 & $\boldsymbol{0.9908}$ & 0.9880\\
(a) - FMNIST  & 0.7978 & 0.8385 & 0.8424 & $\boldsymbol{0.8435}$ \\
(b) - FMNIST & 0.8845 & 0.8776 & 0.8930 & $\boldsymbol{0.8985}$\\
(a) - CIFAR-10  & 0.3226 & 0.3291 & $\boldsymbol{0.4185}$ & 0.3869 \\
(b) - CIFAR-10 & 0.6145 & 0.5217 & 0.6355 & $\boldsymbol{0.6499}$\\
\bottomrule
\end{tabular}
\end{center}
\caption{The accuracy realized by the selected architectures when endowed with ours and state-of-the-art pooling layers.}\label{tab:accuracy}
\label{tab:accuracy}
\end{table}

\paragraph*{Performance.} We tested the architectures presented in~\cref{sec:architectures} on classification of the MNIST, Fashion-MNIST, and CIFAR-10 datasets. For the latter dataset, we adapted the persistence-based and convolutional layer to work with RGB images. Specifically, the persistence-based pooling layer treats channels independently, thus in parallel and without allowing for any interactions among them.

We compared the performance of several pooling layers, namely:
\begin{itemize}
  \item max-pooling,
  \item persistence-based-pooling (ours),
  \item a combination of persistence-based- and max-pooling,
  \item LEAP~\cite{sun2017learning}.
\end{itemize}
The combination between persitence-based- and max-pooling layers is realized by adding the value obtained by max-pooling on a specific patch to the weighted sum defined in~\cref{sec:pfl}. 

We report the results in terms of accuracy in~\cref{tab:accuracy}. The proposed layers outperform max-pooling and LEAP on both datasets. The combination of max- and persistence-based-pooling realizes maximum performance in three out of six experiments. This hints to the complementarity of the two approaches.

\begin{remark}
We combine the persistence-based- and max-pooling simply by adding the maximum intensity value of the patch to the weighted sum defined in~\cref{sec:persistence_based_layers} and described in~\cref{fig:pooling}, panel (c).
\end{remark}

\paragraph*{Explainability}

\begin{figure}[tb]
  \centering
  \includegraphics[width=1\linewidth]{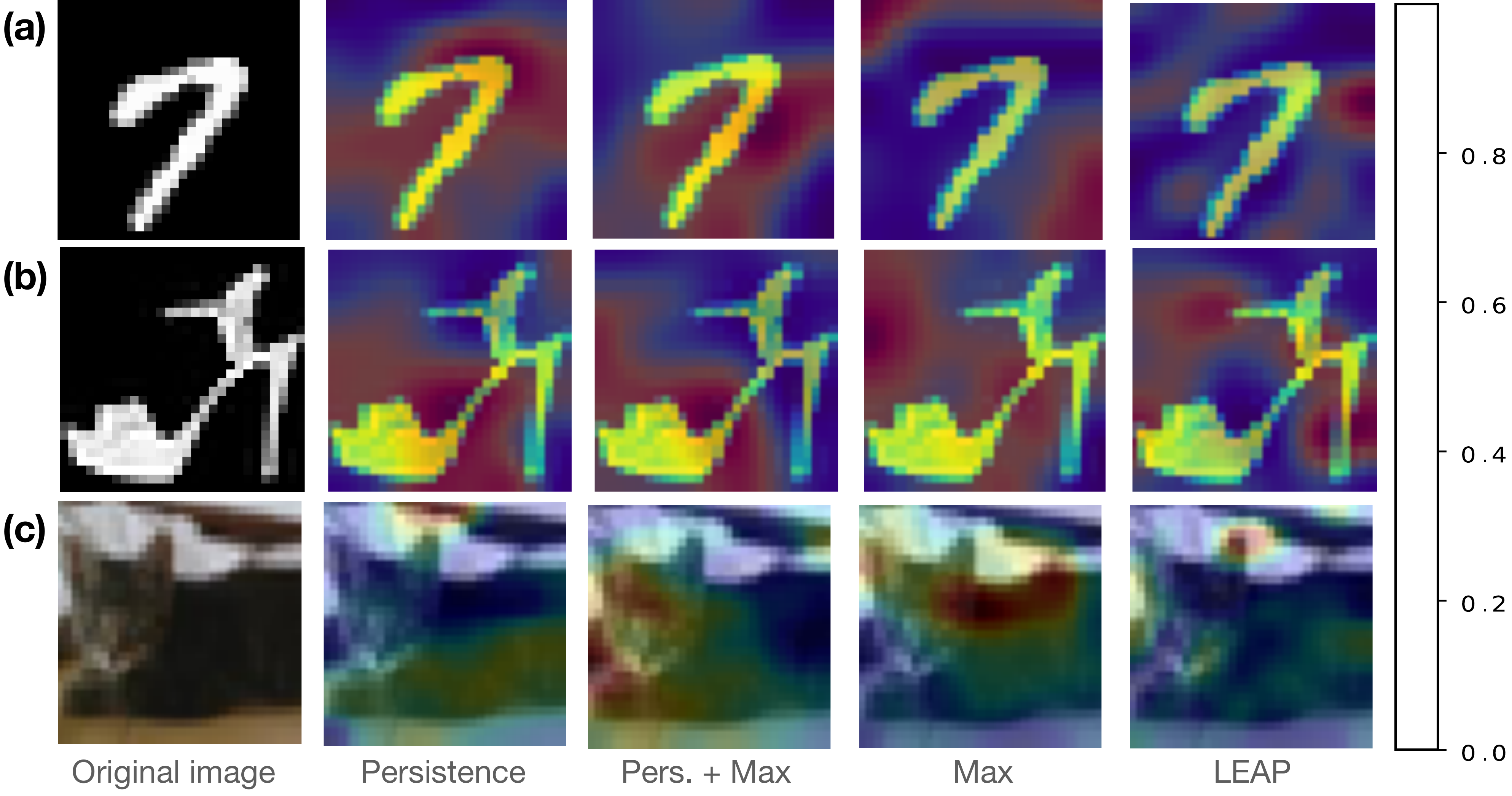}
  \caption{Grad-CAM heatmaps show pixel-wise saliency of features of an image with respect to the neural network's gradients. Each row provides an example from a different dataset: MNIST, Fashion-MNIST, and CIFAR-10, respectively.}
  \label{fig:attention_maps}
\end{figure}

We believe that persistence (topological or generalized) can be of great help in making artificial intelligence explainable, an issue that is luckily gaining more and more attention \cite{holzinger2022explainable,angerschmid2022fairness,cabitza2023quod}
Pooling operators aim to select salient features of their input. We evaluate the similarity across the features detected by the examined pooling layers via  Grad-CAM heatmaps~\cite{selvaraju2017grad}. Some examples are showcased in~\cref{fig:attention_maps}. 

\section{Conclusions}
\label{sec:conclusions}

We work at the crossroad between artificial neural networks, and topological persistence and its generalizations. On the one hand, artificial neural networks can approximate arbitrarily complex functions optimizing an error functions with respect to their input data. On the other hand, the persistence framework provides swift strategies to build easily constrainable analysis pipelines.

We introduce persistence-based layers to pave the road to the design and implementation of hybrid strategies: data-driven and easily constrainable. We base our proposal on the rank-based persistence framework and the concept of steady persistent feature. This foundation endows our layer of important properties such as equivariance and robustness to noise, without requiring auxiliary topological construction that could, \textit{a priori} alter the information originally carried by data of non-topological nature.

Computational experiments on image classification showcase the performance of the proposed persistence-based pooling---a special case of persistence-based layer---against standard pooling operators. Moreover, a pixel-wise saliency analysis through Grad-CAM heatmaps on input images reveals how the proposed pooling layer relies on different features than max-pooling and LEAP.

This approach meshes well with the framework developed in~\cite{bergomi2022neural}, where inputs, outputs, and weights of a neural network layer are expressed as functions on smooth or discrete spaces. There, we showed how several classes of {\em linear} neural network layers can be expressed as a combination of function pullback (from a smaller to a larger space), pointwise multiplication of functions, and integration along fibers. We believe that, by combining the two approaches, it will be possible to define general parametric nonlinear layers, where the architecture is defined by means of the {\em parametric spans} introduced in~\cite{bergomi2022neural}, whereas the geometry-aware nonlinear computation descends from the methods developed here.

\section*{Acknowledgement}
Work performed under the auspices of INdAM-GNSAGA.

\bibliographystyle{elsarticle-num}
\bibliography{Pooling}
\end{document}